# Remote Sensing Image Classification Using Convolutional Neural Network (CNN) and Transfer Learning Techniques

[1]**Mustafa Majeed Abd Zaid**, [1]**Ahmed Abed Mohammed** and [2]**Putra Sumari**

[1]*College of Technical Engineering, Islamic University, Najaf, Iraq*
[2]*School of Computer Science, University Sains Malaysia, Peneng, Malaysia*

**Abstract:** This study investigates the classification of aerial images depicting transmission towers, forests, farmland, and mountains. To complete the classification job, features are extracted from input photos using a Convolutional Neural Network (CNN) architecture. Then, the images are classified using Softmax. To test the model, we ran it for ten epochs using a batch size of 90, the Adam optimizer, and a learning rate of 0.001. Both training and assessment are conducted using a dataset that blends self-collected pictures from Google satellite imagery with the MLRNet dataset. The comprehensive dataset comprises 10,400 images. Our study shows that transfer learning models and MobileNetV2 in particular, work well for landscape categorization. These models are good options for practical use because they strike a good mix between precision and efficiency; our approach achieves results with an overall accuracy of 87% on the built CNN model. Furthermore, we reach even higher accuracies by utilizing the pretrained VGG16 and MobileNetV2 models as a starting point for transfer learning. Specifically, VGG16 achieves an accuracy of 90% and a test loss of 0.298, while MobileNetV2 outperforms both models with an accuracy of 96% and a test loss of 0.119; the results demonstrate the effectiveness of employing transfer learning with MobileNetV2 for classifying transmission towers, forests, farmland, and mountains.

## Introduction

Research in the crucial area of aerial landscape picture categorization has enormous implications for many different fields, including but not limited to biodiversity conservation, urban planning, agriculture, disaster management, climate change studies, military and security activities, and environmental monitoring. To make well-informed decisions and manage resources efficiently, the capacity to correctly categorize and understand aerial photos provides priceless insights and data (Weiss *et al.*, 2020) categorizing aerial pictures is an essential part of environmental monitoring since it allows categorization poses significant hurdles. Transmission towers, woods, agriculture, and mountains are just a few examples of the many land cover categories that conventional picture classification algorithms need help distinguishing between. This restriction impedes attempts to effectively monitor the environment, build cities, and handle disasters, all of which depend on accurate data to make educated decisions. As part of this process, we must track the pace of deforestation, identify instances of illicit logging, and evaluate the state of ecosystems. The systematic cataloging of plant and land cover types may better understand the effects of natural and human-caused changes on ecosystems. Developing plans to safeguard and maintain natural resources is crucial for guaranteeing sustainable development for future generations and this knowledge is essential for that purpose (Haq *et al.*, 2024). CNN and Transfer Learning technology have developed rapidly in recent years; opportunities to improve the precision and efficacy of aerial picture categorization are emerging due to the fast development of deep learning algorithms, especially Convolutional Neural Networks (CNNs) and transfer learning. Yet, strong categorization systems that can use these cutting-edge methods are required due to the complexity and vast amount of aerial imagery produced by contemporary remote sensing technology. Convolutional neural networks are the most commonly used model in deep learning and they have strong self-learning, adaptability, and generalization abilities. Image satellite search has applications in military surveillance, cruising, map remote sensing, traffic control, agricultural farming, and other fields. With the development of an external neural network (CNN) in deep learning, it has been used in research on image classification problems. This study proposes a method for classifying aerial landscape images into four distinct categories: Transmission towers, forests, farmland, and mountains. By leveraging pre-trained models, particularly VGG16 and MobileNetV2, we can accomplish efficient and accurate classification using a Convolutional Neural Network (CNN) architecture. In our Convolutional Neural Network (CNN) design, we have used many convolutional layers, pooling operations and fully connected layers activated with ReLU.

This project aims to develop a high-performance classification system that accurately identifies transmission towers, forests, farmland, and mountains. Additionally, we aim to explore Convolutional Neural Networks (CNNs) performance capabilities for object detection in images. We also aim to demonstrate the efficacy of machine learning approaches, particularly CNNs, in building landscape detection systems. Furthermore, we strive to showcase the improvement in detection quality achieved by employing deep convolutional neural networks. New landscape monitoring and analysis possibilities have emerged due to the fast development of remote sensing technologies, which have generated massive volumes of aerial footage.

Nevertheless, regarding efficient and accurate categorization, these photos' sheer number and complexity pose significant hurdles. Aerial landscapes include a wide variety of complex patterns and traditional picture categorization algorithms are not only sometimes up to the task (Wang *et al.*, 2020), Aerial landscape picture classification utilizing state-of-the-art deep learning techniques: A challenge that this essay seeks to tackle. To that end, we plan to use transfer learning and Convolutional Neural Networks (CNN) to boost the efficiency and precision of categorization. A scalable and reliable method for automatically classifying metropolitan areas, woods, bodies of water, and agricultural fields from high-resolution aerial pictures is the target; in light of these obstacles, this study intends to advance aerial picture categorization by presenting a more workable method; this, in turn, will bolster several applications, including those dealing with environmental monitoring, urban planning, and catastrophe management.

*Related Work*

This article introduces the ARSIC-HHODTL model based on Horse Herd Optimization with Deep Transfer Learning. The model is designed to classify automated remote-sensing images. The primary objective of ARSIC-HHODTL is to develop a method for rapidly and effectively categorizing aerial images using a sophisticated deep-learning model. To begin, we must use bilateral gradient filtering to eliminate noise. The subsequent procedure involves using EfficientNet-B7 on the preprocessed images to extract features. The proposed model is evaluated by assessing accuracy, loss, and MSP metrics. This validation process is carried out to classify and improve the LSTM-based ARSIC-HHODTL model using the HHO technique. The model is tested on several standard datasets. With a performance rate of 94 percent, it surpasses prior techniques, making it very suitable for future applications. The approach exhibits a 5% accuracy improvement compared to previous methods (Rega and Sivakumar, 2024). This study presents a systematic approach for developing a model to categorize aerial scenes using transfer learning. The ReLU-Based Feature Fusion (RBFF) layer selection approach is the foundation of the proposed method. Using the properties of the batch normalization layer in specific blocks of MobileNetV2, RBFF built a model for aerial scene categorization. This model employs MobileNetV2, a framework for single-object picture classification, using feature maps obtained from a pre-trained Convolutional Neural Network (CNN). ReLU activation layers in the associated blocks determine the selection of these blocks. Dimension reduction decreases the number of dimensions in a feature vector, resulting in a space with decreased dimensionality. The reduced feature space trains a nuanced Support Vector Machine (SVM) classifier capable of distinguishing between aerial photos. The newly developed model surpasses previous models in terms of performance on various aerial scene datasets while being cost-effective (Arefeen *et al.*, 2021). The study effort has introduced a novel Multiscale Attention Feature Extraction block (MSAFEB). This characteristic renders it very compatible with seamless integration with other goods or systems since it is designed to be installed and used effortlessly. This block utilizes multiscale convolution at two levels, using skip connections to enable the production of discriminative and salient data at deeper and finer levels. The investigation's performance examination, which included executing the proposed approach on the AID and NWPU benchmark VHR aerial RS picture datasets, demonstrates a consistent and reliable performance with a minimal standard deviation of 0.002. Furthermore, it has an impressive overall recognition rate, up to 94 percent. The accuracies of the NWPU dataset 2009 have standard deviations lower than 95. The average categorization rate in the AID dataset was 85% (Sitaula *et al.*, 2023).

This article proposes a Spectral-Spatial Paralleled Convolutional Neural Network (SSPCNN) for species categorization of forest trees using UAV (uncrewed aerial vehicle) HSI data. In the SSPCNN configuration, a 1-D-CNN is used to learn spectral properties, while a 2-D-CNN is employed to extract spatial information. Experimental findings show that SSPCNN performs competitively compared to other approaches when these characteristics are merged and categorized using a softmax classifier (Liang *et al*., 2020). In this piece, we will look at how to classify aerial images taken by UAVs efficiently for use in disaster response. In it, we examine current methods and provide an Aerial Image Database specifically designed for Emergency Response applications. Using atrous convolutions for multiresolution features, Emergency Net is suggested as a lightweight design for convolutional neural networks. With a speed boost of up to 20 times and an accuracy decrease of less than 1% compared to state-of-the-art models, this lightweight architecture can function effectively on low-power embedded devices (Kyrkou and Theocharides, 2020). This research uses the transfer learning approach. By transferring the weights of pre-trained deep neural networks, this approach applies deep learning models to tiny data, improving recognition accuracy. The research suggests building a UAV dataset using various UAV form architectures to enhance detection and classification capabilities. Models for detection and classification using standard deep convolutional neural networks are tested experimentally. The InceptionV3 model, in particular, benefits greatly from the transfer learning approach, as it attains a recall of 96.48% (Meng and Tia, 2020). A novel model for high-density crowd recognition and classification in aerial photos is presented in this Study. It combines VGG16 with a Kernel Extreme Learning Machine (KELM). Before using the VGG16 technique to extract features, the model goes through preprocessing to enhance the quality of the images. We use the KELM method as a classifier to determine how many people are in the crowd. Through simulations, we can see that the VGG16-KELM technique outperforms the competition. Improving crowd recognition and classification in aerial photos and preventing crowd tragedies during complicated mass events are the goals of the project (Sivachandiran *et al*., 2022).

This study introduces an Optimal Deep Learning Enabled Object Recognition and Classification on Drone Imagery (ODL-ODCDI) method. Objects in drone photos may be identified and classified using this method, which employs ensemble transfer learning. It detects objects using YOLO-v5, a random forest classifier, and a Nadam optimizer. According to the experiments, the ODL-ODCDI technique achieves better results in object recognition and classification on drone photos than other DL models (Adhikari *et al*., 2022). Remote-sensing image scene classification is crucial for various applications, including forest fire monitoring and land-use classification. With the increasing amount of data, researchers have accelerated this process. Advances in computer vision have allowed for the classification of natural images or photographs taken with ordinary cameras. Transfer learning, a technique in many fields, has been successfully applied to natural image classification using convolutional neural network models. The ultimate performance is heavily affected by the hyperparameters that are used to train the models. Models based on smaller remotely sensed datasets perform worse than those trained on bigger, more general datasets of natural images. However, this doesn't diminish the usefulness of transfer learning for scene categorization using distant sensing (Lima and Marfurt, 2019). This study showcases a dataset using a transfer learning technique based on deep learning sub-topics, using the Fire Luminosity Airborne-based Machine Learning Evaluation. The dataset included deep learning methods, including InceptionV3, DenseNet121, ResNet50V2, NAS Net Mobile, and VGG-19, as well as mixed approaches like Support Vector Machine, Random Forest, Bidirectional Long Short-Term Memory and Gated Recurrent Unit. When assessing performance, the DenseNet121 model was 97.95% accurate and the transfer learning model was 99.32% accurate. This method might greatly improve forest fire detection and reaction times (Reis and Turk, 2023). This study proposes a Transfer Learning (TL) methodology. The process uses a classification model trained in the CONUS to determine crops grown in different parts of the world. They used harmonized data from Landat-8 and Sentinel-2 and the study trained on a collection of CONUS and NDVI time series pixels with high confidence. Three test locations were utilized to train and implement Random Forest (RF) classification models: Hengshui in China, Alberta in Canada, and Nebraska in the USA. Utilizing TL with NDVI time series throughout the growth season yielded overall classification accuracies of 97.79, 86.45, and 94.86%. On the other hand, LO could surpass TL in classification accuracy faster. In areas lacking training samples, this research offers new possibilities for crop categorization (Hao *et al*., 2020).

This study introduces landslide detection and classification techniques using Distant Domain Transfer Learning (DDTL). It improves data extraction by introducing scene categorization satellite images and an Attention Mechanism (AM-DDTL). This research compares Convolutional Neural Networks (CNNs), pre-trained models, and DDTL on 177 samples taken from the Longgang study region. Based on the testing findings, DDTL outperforms regular CNN in detection and achieves 94% classification accuracy, which is 7% better than the standard DDTL. When identifying and categorizing possible landslides in various catastrophe zones, the AM-DDTL algorithm performs better than conventional CNN approaches (Qin *et al*., 2021). This study developed a deep land structure model using satellite imagery data from MLRSNet. The model compared three architectures: CNN, ResNet-50, and Inception-v3. The CNN model achieved the highest accuracy of 94.8%, outperforming the other models. The CNN model demonstrated exceptional accuracy, recall, and F1 scores, highlighting its potential in scene understanding and efficient land structure identification from satellite imagery (Abd Zaid *et al.*, 2024).

This article suggests a categorization strategy to enhance traffic management and road safety by combining Convolutional Neural Networks (CNNs) with transfer learning. The system uses a CNN architecture consisting of 24 convolution layers and eight fully connected layers, which is trained using a dataset consisting of 7616 pictures. InceptionV3 performed the best during the validation phase, with a 98.9% accuracy rate. During testing, the suggested CNN model achieved an accuracy of 94.4 % and during validation, it hit 95.1% (Abd Zaid *et al.*, 2025).

## Materials and Methods

Our methodology consists of a set of phases such as dataset description, preprocessing, classification models, and Hyperparameters tunning, as shown in Fig. (1).

### Phase 1: Dataset Description

In this phase, we will explain the dataset in detail. Our model's dataset is a valuable combination of two sources.

MLRSNet (Qi *et al.*, 2020) and self-collected data using maps, as shown in Fig. (2). This comprehensive dataset comprises 10,400 images and encompasses four distinct landscapes: Transmission towers, Forests, Farmlands, and Mountains. These landscapes are classified into four classes, forming the basis of our classification task. By incorporating MLRSNet data and our self-collected data, we aim to create a robust and diverse dataset that accurately represents the real-world scenarios our model will encounter. The MLRSNet data provides a foundational set of images with reliable labels. At the same time, our self-collected data, obtained through extensive mapping efforts, supplements the dataset with additional samples and enhances its diversity.

### Phase 2: Preprocessing

In this phase, we will show a preprocessing set applied to the image before the classification.

### Resizing Image

Essential preprocessing steps include resizing the image while preserving its aspect ratio (Talebi and Milanfar, 2021), stretching it to suit a different size, and changing its dimensions. Every picture in our collection was automatically reduced to 224×224×3 pixels. The raw images gathered were different proportions and sizes than the MLRSNet dataset. This was revealed in a visual inspection of the photographs' aspect ratios. We will not crop the picture. We will crop the image on the left and right sides to create a square when the width exceeds the height. If the width is smaller than the height, we will crop and return the top and bottom portions of the picture as a square. The resulting resized photos will have a resolution of 224×224×3 to ensure that each image is represented by three color channels (red, green, and blue). The samples for each class in the dataset are shown in Fig. (3).

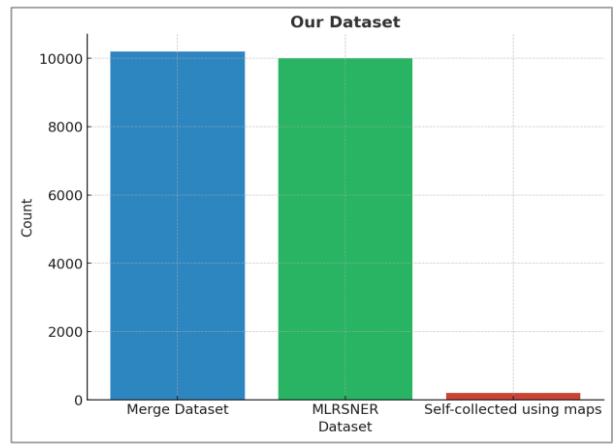

**Fig. 2:** Dataset of study

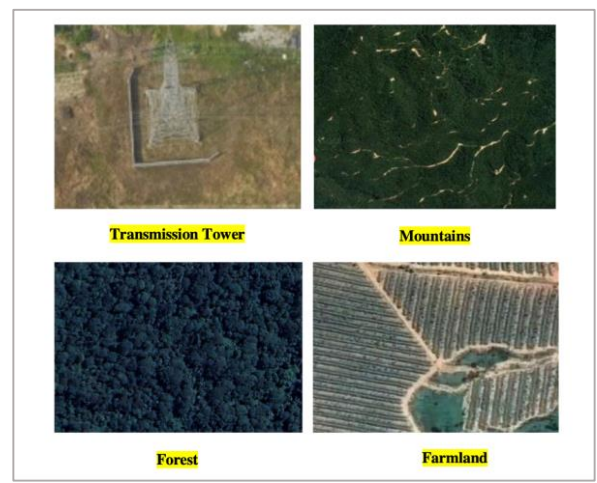

**Fig. 3:** Samples of dataset

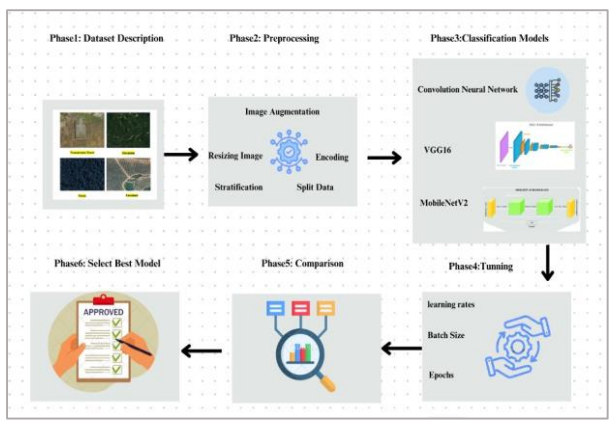

**Fig. 1:** Block diagram of study

*Image Augmentation*

Using image-augmentation techniques, we were able to expand the dataset. One typical approach to enhancing the general-isolation Model's performance is to include data variations via image augmentation (Shorten and Khoshgoftaar, 2019). Specifically, the number of photos via data augmentation.

*Encoding*

Alters the representation of values. Ordinal encoding is another name for label encoding when working with categorical data. Each dataset category is assigned A unique integer value during encoding (Potdar *et al.*, 2017). This method determines the input and filter weights by performing the dot product to convert convolutional filters. Using these methods, the network can examine the input picture and retrieve beneficial characteristics and geographical data (Mohiuddin *et al.*, 2021).

*Split Data*

To ensure practical model training and evaluation (Sugali *et al.*, 2021), we have divided our dataset into three subsets: Sets of training, validation, and testing. 70% of the photos are in the training set, with 15% each in the validation and test sets. Through this partitioning, we can train our model on a substantial subset of the data, refine it using the validation set, and evaluate its efficacy on the separate test set.

*Stratification*

This is a way to guarantee that different data subsets maintain the same distribution of dataset classes. For classification tasks, where biassed models might arise from an unequal class distribution, this was executed during the split to guarantee that each dataset batch had the same number of photos for each class.

*Phase 3: Classification Models*

In this phase, we will present the classification models that we used.

*Convolutional Neural Network (CNN)*

It is a deep learning model developed specifically for handling datasets with a grid pattern, such as photographs. It has several layers, including fully connected, pooling, and convolutional. CNNs use the idea of convolution to their advantage, where small filters are applied to local regions of the input data to extract meaningful features. These features are then progressively learned and aggregated through pooling operations, reducing the spatial dimensions. The learned features are then flattened and fed into fully connected layers for classification or regression (LeCun *et al.*, 2015). CNNs excel at capturing hierarchical patterns in images as the convolutional layers learn to detect low-level features like edges and gradually learn higher-level features. With their ability to automatically learn relevant features, many computer vision tasks, such as picture segmentation, object identification, and classification, have successfully used CNNs.

The architecture of CNN: To successfully categorize pictures with input images with dimensions of 224×224×3, our suggested CNN architecture uses convolutional layers with varying filter sizes, stride values, and pooling operations. A 7×7 convolutional layer adds 64 filters to the model's initialization. To keep the spatial dimensions intact, this layer uses the same padding and a stride of 2 to sample down the feature maps. Next, we have the max pooling layer, decreasing the spatial dimensions while keeping crucial information. It has a pool size of 3×3 and a stride of 2. The same padding is used to retain spatial information in a second convolutional layer with 128 3×3 filters, as seen in Figs. (4-5). After that, the feature maps are down-sampled using an additional max pooling layer. To capture and extract increasingly complex characteristics from the pictures, the following convolutional layers use 256 3×3 filters.

An additional max pooling layer is included after the fourth convolutional layer to reduce the spatial dimensions further and enable the model to concentrate on the crucial characteristics. The following feature maps are flattened to convert the multi-dimensional representations into a one-dimensional vector, which increases the model's capacity. Two completely connected layers, each having 512 neurons activated by a Rectified Linear Unit (ReLU), are then linked to this vector. The ReLU activation function facilitates the model's learning of intricate feature relationships by introducing non-linearity.

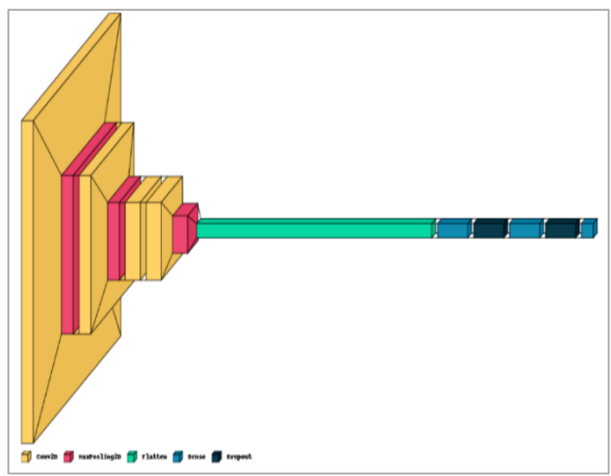

**Fig. 4:** Convolutional neural networks architecture

```
Layer (type)                 Output Shape              Param #
=================================================================
conv2d_44 (Conv2D)           (None, 112, 112, 64)      9472
max_pooling2d_27 (MaxPoolin  (None, 55, 55, 64)        0
g2D)
conv2d_45 (Conv2D)           (None, 55, 55, 128)       73856
max_pooling2d_28 (MaxPoolin  (None, 27, 27, 128)       0
g2D)
conv2d_46 (Conv2D)           (None, 27, 27, 256)       295168
conv2d_47 (Conv2D)           (None, 27, 27, 256)       590080
max_pooling2d_29 (MaxPoolin  (None, 13, 13, 256)       0
g2D)
flatten_9 (Flatten)          (None, 43264)             0
dense_29 (Dense)             (None, 512)               22151680
dropout_18 (Dropout)         (None, 512)               0
dense_30 (Dense)             (None, 512)               262656
dropout_19 (Dropout)         (None, 512)               0
dense_31 (Dense)             (None, 4)                 2052
=================================================================
Total params: 23,384,964
Trainable params: 23,384,964
Non-trainable params: 0
```

**Fig. 5:** Convolutional neural networks diagram

Half-rate dropout layers are added after every fully connected layer to avoid overfitting and enhance generalization. These layers randomly deactivate specific neurons throughout training to improve the model's generalizability to new data. This makes the model less dependent on any one neuron.

Last, four neurons are in the output layer, equal to the number of classes in the classification job. The chance of the input picture belonging to a particular class is represented by probability scores, which are generated by applying the SoftMax activation function to each class. Finding a happy medium between model size and complexity is what this model provides. Adding dropout layers after the fully connected layers improves its capacity to learn complicated features by reducing overfitting that may occur due to complexity. Neurons may be inactivated at random during training via dropout.

Optimizer and Learning Rate: We used the Adam optimizer with different parameters to train our network in our model. The Adam optimizer is famous for deep learning tasks due to its adaptive nature and efficient gradient descent algorithm (Kingma and Ba, 2014). It combines the advantages of two other optimization methods, Ada Grad (Duchi *et al*., 2011) and RMS Prop, to provide effective weight updates during training (Tieleman and Hinton, 2012).

During training, the optimizer modifies the model's weights at a step size determined by the learning rate parameter. It controls the model's convergence and performance. We want to compromise stability and agility during optimization, so we set the learning rate at 0.001. More steady convergence is possible with a lower learning rate, although training may be slower. While a higher learning rate helps expedite training, it also runs the danger of exceeding the ideal answer.

A categorical cross-entropy loss function, appropriate for multi-class classification problems, is used during training. This loss function evaluates how different the actual class labels are from the expected class probability. To minimize the categorical cross-entropy loss, our model gives more weight to the suitable classes and less to the wrong ones.

To train our model, we choose a batch size of 90. This size dictates how many samples are processed before the optimizer updates the weights. Memory and processing performance are both improved by training in batches. In addition, there are ten epochs of training the model, which represents the total number of times the training dataset is traversed.

By measuring the percentage of adequately identified samples, we may evaluate the model's performance during training by looking at the accuracy metric. To keep an eye out for any problems with overfitting or generalization, we also consider the model's performance on a second validation dataset (x val, y val). We maximize the model's performance and get precise categorization using those parameters.

*Transfer Learning*

A pre-trained model is used as a base in transfer learning. However, only the lower layers are typically frozen or kept fixed instead of using the entire model. In contrast, the upper layers are modified or replaced with new classifier layers. This transfer learning approach allows us to benefit from the general knowledge and patterns learned by the base model, which can be applied to various related tasks. The model can improve performance and efficiency in the new application by reusing and adapting the learned features, even with a smaller dataset. Here, we use vgg16 (Simonyan and Zisserman, 2014) and MobileNetV2 (Sandler *et al*., 2018) transfer learning models to compare the actheiracy with our developed CNN model del.

We decided to freeze the layers using the transfer learning model and add a new classifier. This decision was based on the computation needs required to train additional layers of the transfer learning model and considerations related to our dataset size and the model's choices.

VGG16: Thirteen convolutional layers and three fully linked layers comprise the VGG16 architecture. VGG16 employs tiny 3x3 filters all around the network to learn more specific characteristics. Additionally, max-pooling layers are used in the design to decrease spatial dimensions and enhance the model's capacity to deal with fluctuations in object location (Alshammari, 2022). Filters in the convolutional layer increase in size from 64 in the first layer to 512 in the later layers sequentially. When input pictures are stacked using convolutional layers, VGG16 can learn intricate hierarchical representations.

The feature maps are then flattened and sent through three fully connected layers, each with 4096 neurons, after the convolutional layers. These layers provide the classifier and final predictions for the picture classes. By using the rectified linear unit (ReLU) activation function network-wide, non-linearity is brought about.

Many computer vision applications use VGG16 because of its well-known simplicity and efficacy. On the other hand, training VGG16 may be memory and computationally intensive because of all the parameters. This leads to its frequent use as a feature extractor or task-specific transfer learning optimization tool.

MobileNetV2: is a Convolutional Neural Network (CNN) architecture explicitly designed for mobile and resource-constrained devices. It is an evolution of the original MobileNet architecture developed by Google. MobileNetV2 aims to achieve high performance while maintaining efficiency and low computational cost (Gulzar, 2023).

The critical feature of MobileNetV2 is the use of depth-wise separable convolutions. Inverted residuals use a combination of 1x1 and 3x3 convolutions to efficiently capture low-level and high-level features. Linear bottlenecks help preserve the information flow and reduce the model's memory footprint.

An essential aspect of MobileNetV2 is using a technique called width multiplier. This parameter controls the number of filters in each layer, adjusting the model's width based on the available resources. By varying the width multiplier, MobileNetV2 can trade-off between model size and performance, making it adaptable to different devices and applications.

## Results and Discussion

We used our landscape dataset to experiment. On the first pass, we partitioned the dataset into three sections: Training (70%), validation (15%), and testing (15%). Iteratively, training and validation take place simultaneously. To train a reliable model, we monitored the impact of various factors and fine-tuned them accordingly. This model is implemented using the Python library" Keras" on a laptop with an AMD RADEON RYZEN 5, 16 GB of RAM, and 400 GB of disk space, using Jupiter Notebook.

### Results

We will show all the results that we obtained from our experiment.

### Effects of Hyper-Parameters

Effect of batch size: The batch size is a hyperparameter that determines the number of samples the model processes in each training iteration. It impacts the training time and the model's ability to generalize to unseen data (Goodfellow *et al*., 2016). Training with larger batches generally results in faster training times since the model processes more samples in each iteration. However, this may require more memory to store the activations and gradients, which can be a limitation for resource-constrained systems (Yoshua, 2013).

To show the effect of the batch size on our model, we chose different values of batch size 90, 50, and 15, conserving the same learning rate and epoch on the first model we built from scratch. The results are shown in Table (1).

Effect of number of epochs: When training a machine learning model, the amount of epochs is measured by how many times the entire dataset is transferred forward and backward. Many iterations, each handling a separate data batch, make up an epoch. The frequency with which the model updates its weights using the training data is controlled by the hyperparameter known as the number of epochs. The model can pick up on more intricate patterns as the training time increases, leading to better performance. However, overfitting may happen if the model needs to be more specific to the data used for training. Underfitting, in which the model misses important data patterns, may occur if training for insufficient epochs. Considerations such as dataset size, model complexity, and computing resources dictate the ideal number of epochs. Using the same batch size and learning rate on the VGG16 transfer learning model, we tested three alternative epoch values (10, 4, 2) to demonstrate his effects. You can see the results in Table (2).

**Table 1:** Batch effect on the model accuracy

| Batch size | Final Training accuracy | Final Validation accuracy |
|---|---|---|
| 90 | 96.6% | 84.1% |
| 50 | 91.1% | 82.7% |
| 15 | 91.6% | 82.1% |

**Table 2:** Epochs number effect on the model accuracy

| Epoch number | Final training accuracy | Final validation accuracy |
| --- | --- | --- |
| 10 | 91.3% | 87.6% |
| 4 | 90.0% | 89.2% |
| 2 | 90.8% | 88.2% |

We can observe that modifying the batch size affects the model's accuracy, depending on the dataset and the specific model used. In our case, a batch size of 90 resulted in the highest accuracy.

Changing the number of epochs resulted in varying levels of accuracy. In this case, the model with an epoch size of 10 achieved the highest accuracy.

Effect of learning rates: One of the hyperparameters used to train a model is the learning rate, which controls the increment by which the parameters are updated. It can significantly affect the model's convergence and performance as a critical training component. More comprehensive parameter updates are possible with a greater learning rate, which may speed up convergence but increase the likelihood of overshooting the ideal solution. In contrast, more minor updates are produced by a lower learning rate, which can lead to slower convergence but may help the model to converge to a more precise solution. Selecting an appropriate learning rate involves balancing convergence speed and accuracy. To display the effect of the learning rate, we selected three different learning values, 0.01,0.001,0.0001, and applied them to our last model, MobileNetV2, conserving the same batch size and epoch. The results are shown in Table (3).

According to Table (3) the best accuracy for this model was achieved by applying a learning rate of 0.0001.

## Models Validation Accuracy and Loss Comparison

In Fig. (6) we can see that our proposed model performs less effectively compared to the two-transfer learning models. Among the models evaluated, MobileNetV2 exhibited the highest performance.

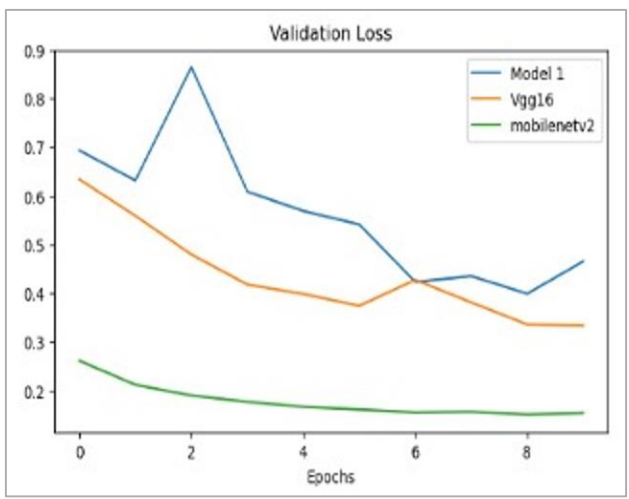

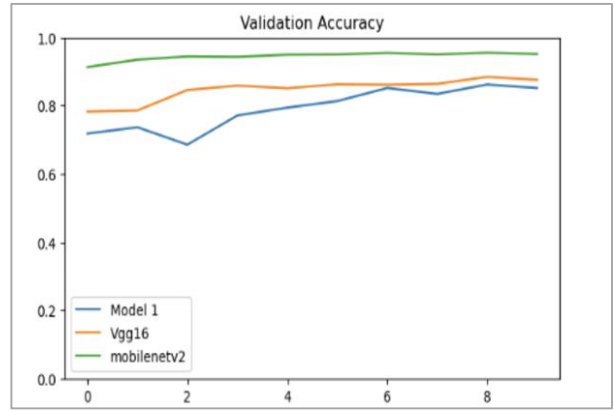

**Fig. 6:** Validation accuracy and loss comparison

**Table 3:** Learning rate effect on the model accuracy

| Learning rate | Final training accuracy | Final validation accuracy |
| --- | --- | --- |
| 0.01 | 97.7% | 95.1% |
| 0.001 | 00.0% | 95.6% |
| 0.0001 | 99.5% | 95.5% |

**Table 4:** Comparison of different models' accuracy

| Model | Test accuracy | Test loss |
| --- | --- | --- |
| Proposed model | 87.2% | 0.363988041 |
| MobileNetv2 | 96.1% | 0.119421191 |
| Vgg16 | 90.6% | 0.297783434 |

We can compare the three models' validation accuracy and loss using the fixed value of 90 batch size, ten epochs, and a learning rate of 0.001.

## Discussion

For the testing step, we utilized a batch size of 90, ran the model for ten epochs, used the Adam optimizer, and set the learning rate to 0.001. Table (4) shows that the proposed model achieved a test accuracy of 87.2% and a test loss of 0.364. Although the accuracy is satisfactory, there is room for improvement, particularly when comparing it to MobileNetV2 and VGG16.

MobileNetv2 achieved a significantly higher test accuracy of 96.2% and a lower test loss of 0.119. This indicates that MobileNetv2 performs better than the proposed model in terms of accuracy and generalization. With its efficient architecture,

MobileNetv2 is suitable for our specific task. VGG16 achieved a test accuracy of 90.6% and a test loss of 0.298. While VGG16 performs reasonably well, accuracy needs to improve compared to MobileNetV2. The VGG16 architecture may be too complex for our specific task, but it is worth considering.

Overall, MobileNetV2 outperforms both the proposed model and VGG16 in terms of test accuracy and loss. It strikes a good balance between accuracy and efficiency, making it a suitable choice. The proposed model demonstrates reasonable performance but has room for improvement. While VGG16 may not be as accurate as MobileNetV2, it can still be valuable in more complex feature extraction scenarios.

## Conclusion and Future Work

In conclusion, this study uses deep learning techniques to classify aerial landscape images by combining MLRSNet and self-collected data. Our proposed Convolutional Neural Network (CNN) architecture and transfer learning models VGG16 and MobileNetV2 were evaluated for landscape classification. Transfer learning involves utilizing a pre-trained model and adapting it for a specific task by modifying or replacing the upper layers. VGG16, known for its simplicity and effectiveness, performed reasonably well, while MobileNetV2, designed for efficiency on resource-constrained devices, outperformed both the proposed model and VGG16 regarding accuracy and loss.

Our findings highlight the effectiveness of transfer learning models, particularly MobileNetV2, for landscape classification. These models balance accuracy and efficiency, making them valuable choices for real-world applications. The proposed model demonstrates reasonable performance and can be further optimized to enhance accuracy. Further research and experimentation can focus on improving the proposed model or exploring other transfer learning architectures tailored to specific landscape classification tasks; adjusting hyperparameters and layers can lead to improved accuracy and performance.

## Acknowledgment

We appreciate the efforts of the Editorial team in reviewing and editing this study. Thank you to the Publisher for the support in the publication of this article.

## Funding Information

This study was funded by the Islamic University in Najaf, Iraq.

## Author's Contributions

**Mustafa Majeed Abd Zaid:** Wrote related works, and also conceived and designed the block daigram of methodology.

**Ahmed Abed Mohammed:** Wrote introduction and some methodology and chapter four.

**Putra Sumari:** Supervisor and contributes to proof the article and do experiments.

## Ethics

This article is original and has never been published before. All Authors have reviewed and approved the piece and the corresponding author states that there are no ethical considerations.